# Real-Time Capable Micro-Doppler Signature Decomposition of Walking Human Limbs

Sherif Abdulatif*, Fady Aziz*, Bernhard Kleiner, Urs Schneider
Department of Biomechatronic System, Fraunhofer Institute for Manufacturing Engineering and Automation IPA
Nobelstr. 12, 70569 Stuttgart, Germany
Email: {sherif.abdulatif, fady.aziz, bernhard.kleiner, urs.schneider}@ipa.fraunhofer.de
* These authors contributed to this work equally.

*Abstract*—Unique micro-Doppler signature ($\mu$-D) of a human body motion can be analyzed as the superposition of different body parts $\mu$-D signatures. Extraction of human limbs $\mu$-D signatures in real-time can be used to detect, classify and track human motion especially for safety application. In this paper, two methods are combined to simulate $\mu$-D signatures of a walking human. Furthermore, a novel limbs $\mu$-D signature time independent decomposition feasibility study is presented based on features as $\mu$-D signatures and range profiles also known as micro-Range ($\mu$-R). Walking human body parts can be divided into four classes (base, arms, legs, feet) and a decision tree classifier is used. Validation is done and the classifier is able to decompose $\mu$-D signatures of limbs from a walking human signature on real-time basis.

## I. INTRODUCTION

A human walk can be analyzed as a human bulk translational motion represented in non-swinging body parts as torso and head. In addition to, the micro-motions of the swinging limbs as legs and arms. The micro-Doppler signature ($\mu$-D) of the limbs can be described as modulations on the returned radar signal due to limbs micro-motions [1]. The resulting $\mu$-D signature is proved to be the superposition of the limbs with the torso $\mu$-D signatures [2]. Thus, the decomposition of $\mu$-D signatures from a superimposed signal into different components corresponding to the motion of each body part is necessary. Some applications require real-time classification of human motion in industrial production to ensure a quick response in unsafe situations such as stopping the machine or warning the worker in less critical cases.

Raj and Chen [3], [4] introduced two $\mu$-D signatures decomposition techniques based on modelling of the Time-Frequency (TF) representation. The first proposed technique is a parametric method to decompose body parts signatures. In this technique, a Gaussian g-Snake model is employed to obtain $\mu$-D signatures from the smoothed manifold curves of moving body parts as a prior knowledge. These curves are represented as functions of some parameters of interest where a steepest descent algorithm is adopted to minimize the mean-square error and estimate such parameters. The second proposed technique is based on non-parametric method where no prior knowledge is involved and peak tracking algorithm substitutes the prior knowledge in the previous method. These proposed algorithms needs a time window of at least half human gait cycle. Thus, this technique can not be employed in a real-time safety application.

This paper considers a walking human scenario where the classification of velocities and ranges of different limbs is applied in real-time. A joint TF analysis can be used to visualize the $\mu$-D signatures of a walking human. A commonly used technique for TF analysis is Short-Time Fourier Transform (STFT), where the spectrogram function $S(t, f)$ is defined as the logarithmic square magnitude of the STFT of the data $x(t)$ with a Gaussian window function $w(t)$ [1], as follows:

$$S(t,f) = \left| \sum_{n=-\infty}^{\infty} w(n)x(t-n)e^{-j2\pi fn} \right|^2 \quad (1)$$

The range and velocity are jointly used in our approach for limbs decomposition. One common approach is used to map corresponding velocities and ranges in [5], where a Discrete Fourier Transform (DFT) of the signal can be applied in two dimensions. In this technique, the data are collected over $N_p$ chirps and each chirp has $N_s$ samples to finally construct a matrix $\boldsymbol{D} \in \mathbb{C}^{N_s \times N_p}$. A two dimensional DFT is applied to the matrix $\boldsymbol{D}$ that can be interpreted as a range-velocity plane. Therefore, the dominant frequency during one chirp will represent the range frequency ($f_r$) and the Doppler frequency ($f_d$) can be interpreted from multiple ramps. Accordingly, a joint analysis to extend the velocity $\mu$-D signatures with a micro-range ($\mu$-R) dimension can be developed [6], [7]. This technique can then be deployed to construct a three dimensional signature ($t - v - r$).

In this paper, we propose a machine learning based approach for human limbs decomposition. The main features involved in our proposed real-time classification are the $\mu$-D velocities and $\mu$-R of human body parts during walking.

The remainder of the paper is organized as follows: Section II introduces the simulated $\mu$-D signature techniques used, Section III describes the joint range-Doppler mapping analysis, while Section IV describes the classification approach used. The results and comparisons of limbs classification will be presented in Section V. The paper is concluded and some future work is presented in Section VI.

## II. SIMULATIONS

### A. Global Human Walking Model

Boulic and Thalmann [8] proposed a walking model, named as "*Global Human Walking Model*", based on empirical mathematical method for translation and rotation of each body part. In [9], it is proved that human body parts can be decomposed into ellipsoids that have smoother reflections than cylindrical parts representation. From [1], the human body is divided into 16 parts and each part can be used to form a 3D ellipsoid. Therefore, backscattered Radar Cross Section (RCS) of human body segments ($\sigma$) can be approximated as a function of the ellipsoid radii lengths $(a, b, c)$ in $(x, y, z)$ directions respectively, incident aspect angle ($\theta$) and azimuth angle ($\phi$) as shown in Eq. 2. The signal model presented in [9] is used to represent the received ramp by the radar at each time instance ($i$) from each body segment. In Eq. 3, the signal model ($A_i$) is represented as a complex value with an amplitude of $\sqrt{\sigma_i}$ and a phase function of the distance ($d$) of each body segment with respect to the radar and the wavelength ($\lambda$).

$$\sigma = \frac{\pi a^2 b^2 c^2}{\left(a^2 \sin^2 \theta \cos^2 \phi + b^2 \sin^2 \theta \sin^2 \phi + c^2 \cos^2 \theta\right)^2} \quad (2)$$

$$A_i = \sqrt{\sigma_i} e^{\frac{-j4\pi d_i}{\lambda}} \quad (3)$$

In the proposed simulation, a radar with a carrier frequency $f_c$ = 25 GHz is placed at a height of 1 m and at a distance of 10 m from the human model. The simulated signal bandwidth $B$ = 2 GHz gives a range resolution $R_{res}$ = 7.5 cm based on the relation $R_{res} = c/2B$, where c is the speed of light. The human is simulated with a height of 1.75 m to walk towards the radar with a relative velocity of 1 m/s according to the model derived in [1]. The received radar ramps are super-positioned from different body segments for the period of interest. The range data are included in the amplitude values derived from the RCS approximation method. Thus, the simulated $\mu$-R signature at each time instance $i$ is estimated as the absolute values of the super-positioned ramps for the 16 different body segments as presented in Eq. 4. As shown in Fig. 1, the swinging limbs effects is observed together with a regression behavior due to walking towards the radar. While the $\mu$-D signature can be visualized using the STFT analysis on the super-positioned ramps of different body parts as shown in Fig. 2.

$$R_i = \sum_{s=1}^{s=16} |A_{s,i}| \quad (4)$$

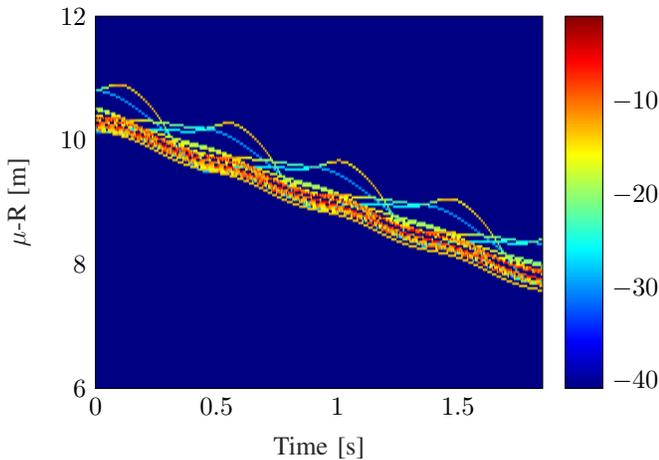

Fig. 1: $\mu$-R signature based on Global Human Walking Model.

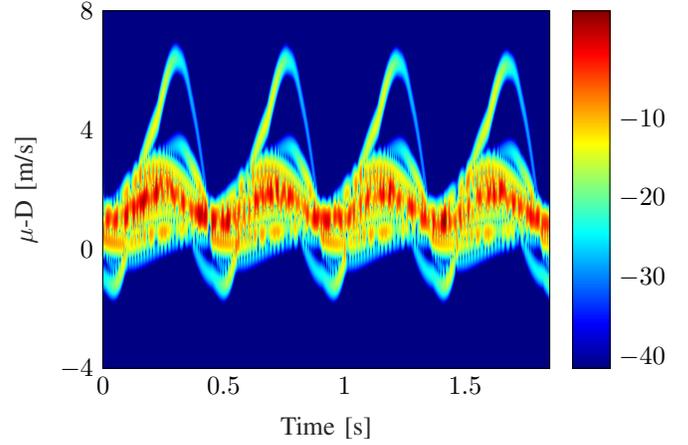

Fig. 2: $\mu$-D signature based on Global Human Walking Model.

### B. Infrared Motion-Capture Combined Model

This model is based on optical motion capture data using the Qualisys[TM] System [10] with 10 infrared cameras placed in a room (5 on each side) to capture infrared-reflective markers on a subject. These markers positions are measured as ground truth motion trajectories of different body components. In our proposed model, 17 markers are placed on a subject according to the Global Human Walking Model. From the velocity curves of some selected body parts shown in Fig. 3, the maximum velocity component is as expected in the toe (foot part) and it reaches a value of $v_{max}$ = 4.5 m/s. Given the subject under test is walking, therefore the relative velocity ($V_{WR}$) is between 1 m/s − 1.5 m/s. As already observed that the maximum swinging velocity is in feet and can reach up to $4 \times V_{WR} \approx 6$ m/s.

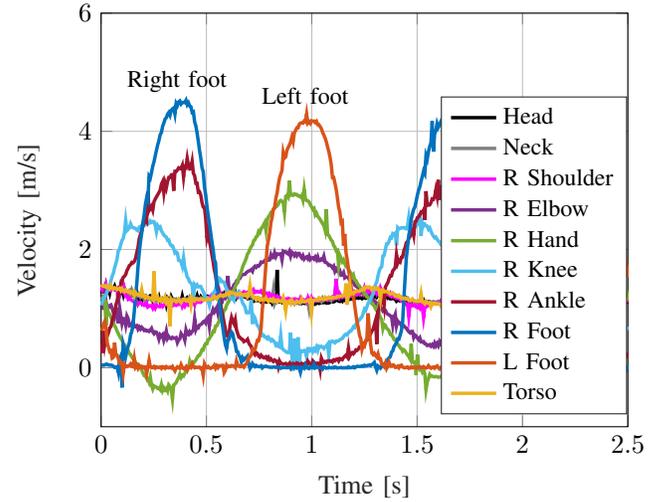

Fig. 3: Velocity curves measured with Qualisys[TM].

From this maximum velocity (6 m/s), the corresponding chirp duration is derived as $T_p$ = 0.6 ms based on Eq.5. Based on Eq. 6, to get a velocity resolution ($v_{res}$ = 0.1 m/s) with chirp repetition duration ($T_p$ = 0.5 ms), then the minimum number of chirps per measurement ($N_p$) to the next multiple of 2 is 64 samples. Both relations for $T_p$ and $N_p$ are described in [5].

$$T_p = \frac{c}{4f_c v_{max}} \quad (5)$$

$$N_p = \frac{c}{2f_c T_p v_{res}} \quad (6)$$

The Global Human Walking model is substituted with real captured position data acquired from the subject reflective markers. However, RCS backscattered ellipsoid approximation proposed previously is still used to approximate different body parts, but with real dimensions of the moving subject limbs. The location of the 25 GHz radar is simulated at a distance of 5 m and a height of 1 m with the same bandwidth of 2 GHz, thus a range resolution of 7.5 cm is achieved. The real relative velocity and height of the subject are also extracted from the data as 1.4 m/s and 1.7 m, respectively. The backscattered RCS is then used to evaluate the $\mu$-R and $\mu$-D signatures of the proposed "*Combined Model*" shown in Fig. 4 and Fig. 5, respectively.

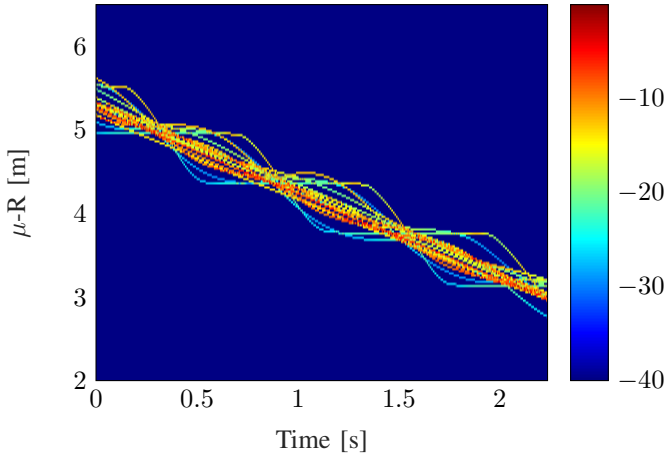

Fig. 4: $\mu$-R signature based on Combined Model.

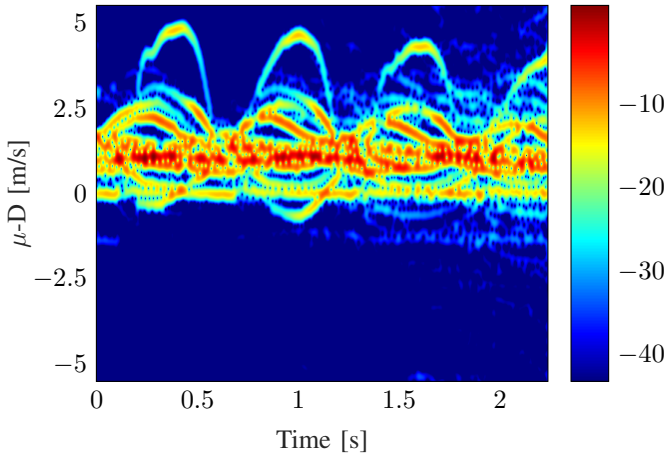

Fig. 5: $\mu$-D signature based on Combined Model.

### III. JOINT $\mu$-D AND $\mu$-R MAPPING

#### A. Range-Velocity Mapping

The procedure for range-velocity mapping presented in [5] is based on forming different data matrices $D$, as shown in Fig. 6. The technique is based on applying FFT analysis on both dimensions in which the range data is estimated within each chirp $N_s$, while the velocity data is estimated across consecutive chirps $N_p$. Thus, each matrix after FFT is analyzed as a range-velocity map. However, in our proposed signal model, the FFT is only needed across the second dimension as the range data are already included in the amplitude values as explained in [9].

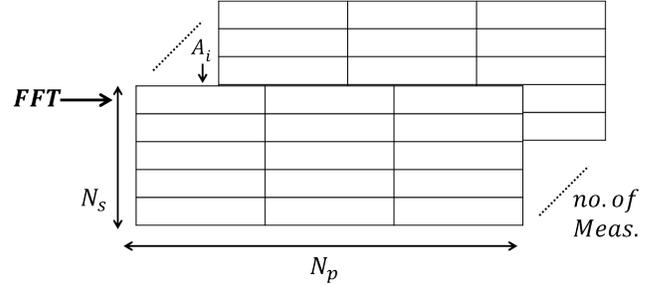

Fig. 6: Range-Doppler Mapping Procedure

#### B. Power Removal

The colored time-frequency representation of the $\mu$-D signature in Fig. 5 is used to indicate the received power by the radar due to the movement of different body parts. However, the power aspect should not be used for the limbs decomposition process. One reason is that the received $\mu$-D value at any time instance is the superposition of reflected signals due to different body parts. Moreover, in real radar measurements the changing human position with respect to the radar will lead to power fading effect. Thus, power will not be included in the classification process.

The power is removed in two steps. First, the noise effect is reduced and the difference between the low and the high $\mu$-D values is reduced by using Gamma-transformation technique described in [11]. Then, the important $\mu$-D values are extracted using the Otsu's thresholding technique. The Gamma-Transformation is an intensity mapping technique that is chosen for its simplicity. The relation between input values $S_{in}$ and mapped values $S_{out}$ is based on the tunning factor $\gamma$ as follows:

$$S_{out} = S_{in}^{\gamma} \quad (7)$$

The selection of $\gamma$ depends on the desired mapping either $\gamma \leq 1$, $\gamma = 1$ or $\gamma \geq 1$. Therefore, $\gamma = 0.65$ is selected in this paper for the $\mu$-D analysis. Then, the Otsu's thresholding method described in [12] is used for extracting the received $\mu$-D values from the captured spectrogram. This technique is based on defining a varying threshold between the desired two classes to be separated by minimizing the summation of the weighted variance of each class which is given as follows:

$$\sigma^2 = Q_1 \sigma_1^2 + Q_2 \sigma_2^2, \quad (8)$$

where, $Q_{1,2}$, $\sigma_{1,2}^2$ are the class probability density and the class variance respectively. The power removal effect is shown in Fig. 7 for 2 different time instances. The power removal preserves the $\mu$-R and the $\mu$-D performance as can be shown in Fig. 8. The regression due to the changing position of the walking human is visible in the range dimension. The swinging effects of the different limbs are visible on both the range and the velocity dimensions.

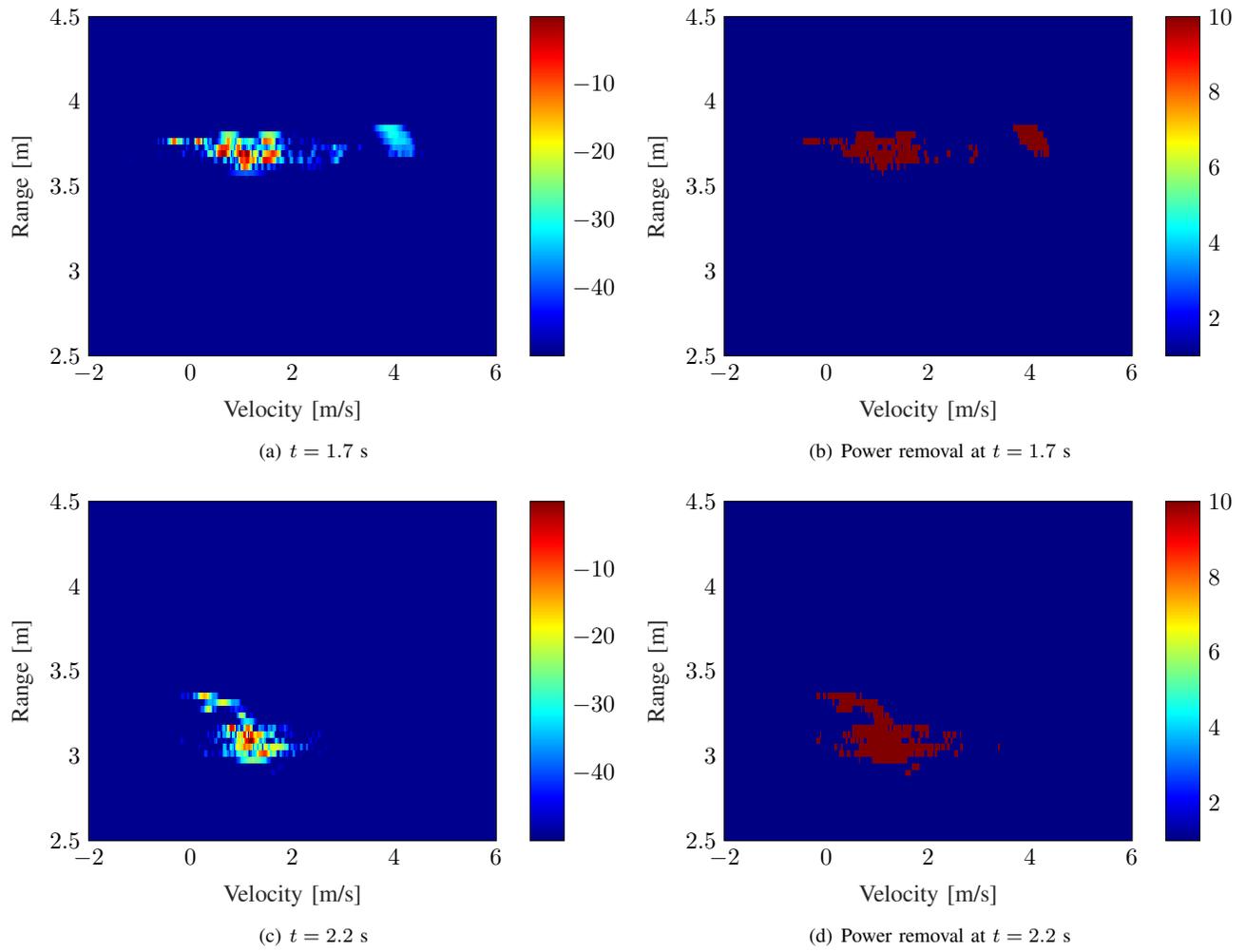

Fig. 7: Mapped $\mu$-D and $\mu$-R signatures.

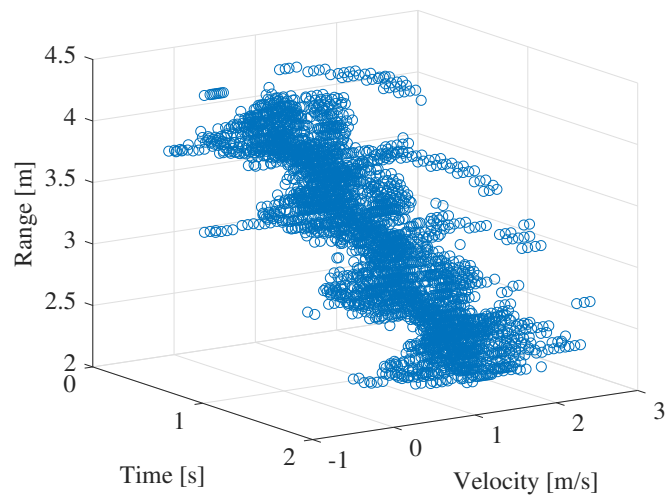

Fig. 8: Scatter plot of the mapped $\mu$-D and $\mu$-R signatures within the walking gait cycle.

## IV. CLASSIFICATION

In this section, the data from the previously stated *Combined Model* will be decomposed by means of ($\mu$-D and $\mu$-R) features into different classes, where each class combines some body parts. As a preprocessing step, mean removal is applied on the $\mu$-R data in each time step to extract only the swing effect without linear regression behavior shown in Fig. 4. Our proposed approach is time independent, so features from the whole experiment period are used to train the classifier.

Due to the body symmetry, the right and left corresponding body parts will have very similar $\mu$-D and $\mu$-R signatures. This will lead to undetermined differentiation of left and right body parts based on $\mu$-D and $\mu$-R only. As shown in Fig. 3, the left and right feet differs by half a gait cycle, which indicates that the velocity phase shifts are the only difference between left and right body parts. However, in a real case scenario, the starting body side of the motion is unknown. Thus, identifying each side alone is difficult without prior knowledge. Moreover, in the previous decomposition techniques, the first moving side is considered as a prior knowledge and then a side flip is applied every half a cycle [3], [4]. Accordingly, the left and right body parts will be considered in a joint class.

Furthermore, the parts of minimum velocity and swings such as the torso, head and neck will be combined together in one class (base). The elbows, upper and lower arms are combined together in one class (arms). For the lower body parts, feet $\mu$-D are relatively higher than the remaining leg parts. Thus, feet can be considered as a separate class as depicted from Fig. 3. Finally, the legs class combines the knee, lower and upper legs. Accordingly, the problem can now be simplified to a four class problem (Base, Arms, Legs and Feet) yielding the $\mu$-D and $\mu$-R scatter plots, shown in Fig. 9.

remaining (hidden) 25% of the data are used for validation. A decision tree classifier is trained and used to get the confusion matrix in Table I. The legs and arms classes are confused at some instants as they experience an overlap in $\mu$-D and $\mu$-R plane shown in Fig. 9. While, the feet get confused with the legs and arms classes with equal probabilities. The base has the least performance because it has only low $\mu$-D and $\mu$-R components, which can also be found in legs and arms classes.

Table I: Limbs decomposition confusion matrix.

| True/Predicted | Arms | Feet | Legs | Base |
|---|---|---|---|---|
| Arms | 65% | 1% | 32% | 2% |
| Feet | 14% | 73% | 13% | 0% |
| Legs | 19% | 1% | 79% | 1% |
| Base | 37% | 0% | 37% | 26% |

New labeled data are collected based on the *Combined Model* as shown in Fig. 11 to validate the real-time decomposition. At each time instant, super-positioned $\mu$-D and $\mu$-R features are extracted and mean removal is applied on $\mu$-R to remove the regression effect. The previously trained classifier is used to label each feature vector.

Multiple samples in the $\mu$-D and $\mu$-R plane can have the same class label at the same time instant. This is expected as each class consists of different swinging effects. For example, the arms class combines the hand which has higher swinging effects than the shoulder. Thus, we propose using the maximum and minimum $\mu$-D features to get an envelope detection behavior for each body part as shown in Fig. 10. Due to the availability of some outliers in the decomposed curves, the median filtering is presented in the next subsection to enhance the decomposed final curves.

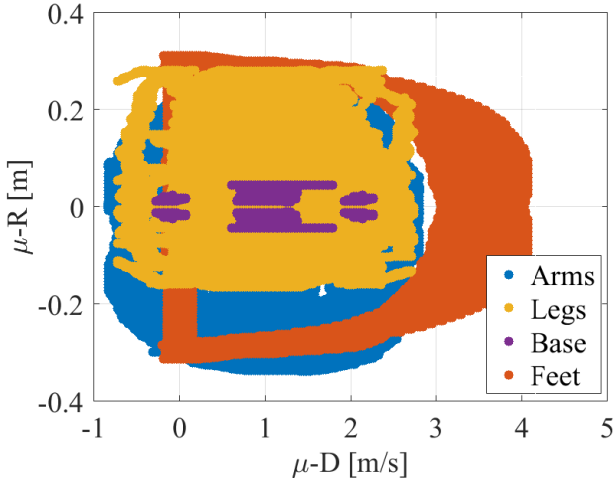

Fig. 9: Combined Model $\mu$-D and mean-free $\mu$-R features.

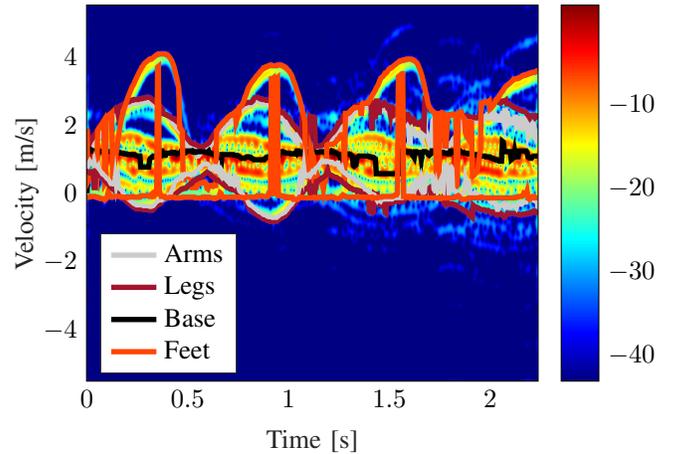

Fig. 10: Decomposed $\mu$-D signatures from Combined Model.

### B. Outliers Removal

The Median Filtering technique of order 9 is chosen for outliers detection and removal. Given the time taken for one range-Doppler map is about 0.03 s. Thus, the median filter will need to wait for a buffer of 4 consecutive instances ($\approx 0.1$ s). This reaction time will preserve the real-time performance; however, it can be improved later using recursive Kalman filter.

Applying the outliers removal technique shows an enhancement in the decomposed curves as shown in Fig. 12. The

## V. RESULTS

### A. Limbs Decomposition Analysis

Based on the *Combined Model*, features from different subjects of different heights and genders are collected and labeled before classification for training and validation over sufficient time span. The data set of about one million time instances is then divided into 75% training data and the decomposed curves show agreement with the labeled data in Fig. 11. Based on this approach, if a half gait cycle duration can be estimated by monitoring the cycle behavior in real-time, then sides (right,left) differentiation can be accomplished by switching between the minimum and maximum envelopes every half a cycle duration. However, labeling each side as right or left is still not possible without prior knowledge. Another improvement can be achieved by further classification of same class points into subclasses to identify subparts as shoulder, elbow, knee, torso.

It can be observed that feet are fully identified at instants of high $\mu$-D values. Outliers are seen at low feet $\mu$-D values shared with other parts which were removed using the median filtering. During walking, the hand and ankle components have similar $\mu$-D velocities. Moreover, both parts have the highest $\mu$-D values in the arms and legs classes, respectively. Thus, when the maximum or minimum $\mu$-D is chosen to represent each class at each time instant, the arms and legs curves give an overlapping behavior with a slight phase shift. From Fig. 9, the base misclassification rate is high due to low $\mu$-D values in legs and arms. However, the base values in Fig. 12 are still in the acceptable low $\mu$-D area indicating the relative human velocity as they get confused with low $\mu$-D values only.

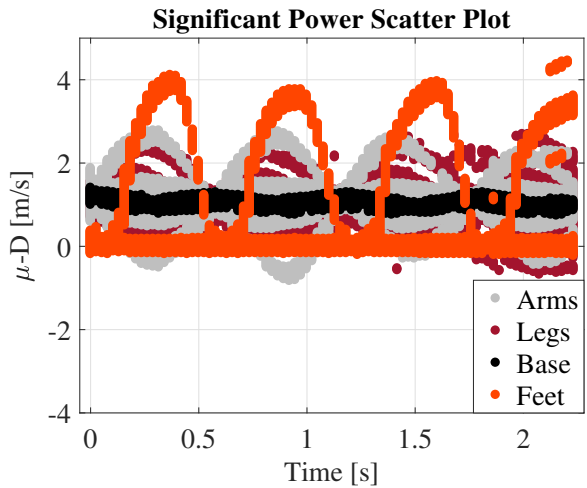

Fig. 11: Labeled $\mu$-D signatures from Combined Model.

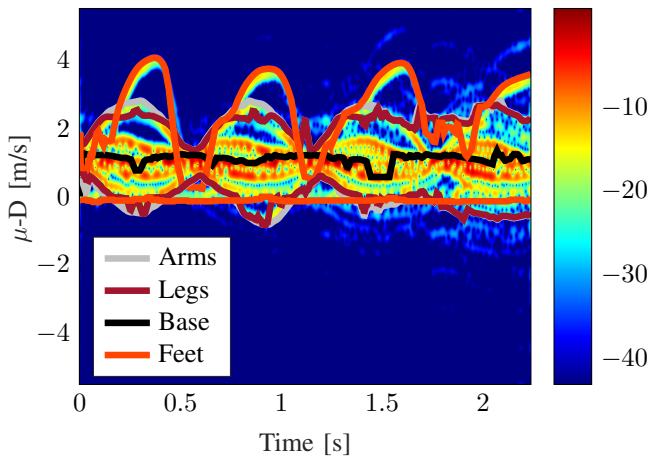

Fig. 12: Decomposed $\mu$-D signatures from Combined Model.

## VI. CONCLUSION

This paper presents a novel real-time capable $\mu$-D limbs decomposition technique of a walking human. Two simulations are combined to model a walking human target towards a radar. The first simulation is based on a mathematical empirical model. In the second simulation, infrared cameras are used to track positions of infrared-reflective markers placed on a real moving subject. Mapped power-free $\mu$-D and mean-free $\mu$-R are evaluated as more significant classification features to be independent from a specific walking target range scenario. The body is then divided into four main classes (base, arms, legs and feet). Body parts are combined in these four classes based on the relevancy of each part to the class. The $\mu$-D and $\mu$-R are extracted from the used model over time and then used as features for a decision tree classifier. The collected features are used to train the decision tree classifier and new data set are generated for validation. An outliers removal algorithm is implemented to improve the true classification rates of the classifier.

The work presented in this paper is a feasibility study for limbs decomposition using machine learning based approach. The presented methodology included a walking human towards the radar. In this case, the walking direction is aligned with the radar on the same axis. However, for future work other possible scenarios can be also included. One scenario will be moving away from the radar and this will reverse the $\mu$-D signature sign. This issue can be addressed by detecting the direction of motion and classifying accordingly. Other scenarios will be moving parallel to the radar radial axis or tangential to the radar. These cases will be addressed by normalization of the $\mu$-D feature in the classification procedure. Furthermore, a validation of the presented algorithm will be done on real radar measurements.

## REFERENCES

[1] V. C. Chen, *The micro-Doppler effect in radar*. Artech House, 2011.
[2] J. L. Geisheimer, E. F. Greneker III, and W. S. Marshall, "High-resolution doppler model of the human gait," pp. 8–18, 2002. [Online]. Available: http://dx.doi.org/10.1117/12.488286
[3] R. G. Raj, V. C. Chen, and R. Lipps, "Analysis of radar dismount signatures via non-parametric and parametric methods," in *IEEE Radar Conference*, 2009, pp. 1–6.
[4] R. Raj, V. Chen, and R. Lipps, "Analysis of radar human gait signatures," *IET Signal Processing*, vol. 4, no. 3, pp. 234–244, 2010.
[5] B. J. Lipa and D. E. Barrick, "Fmcw signal processing," *FMCW signal processing report for Mirage Systems*, 1990.
[6] O. R. Fogle and B. D. Rigling, "Micro-range/micro-doppler decomposition of human radar signatures," *IEEE Transactions on Aerospace and Electronic Systems*, vol. 48, no. 4, pp. 3058–3072, 2012.
[7] R. Rytel-Andrianik, P. Samczynski, D. Gromek, M. Wielgo, J. Drozdowicz, and M. Malanowski, "Micro-range, micro-doppler joint analysis of pedestrian radar echo," in *Signal Processing Symposium (SPSympo)*, 2015, pp. 1–4.
[8] R. Boulic, N. M. Thalmann, and D. Thalmann, "A global human walking model with real-time kinematic personification," *The visual computer*, vol. 6, no. 6, pp. 344–358, 1990.
[9] P. van Dorp, "Human walking estimation with radar," *IEE Proceedings-Radar, Sonar and Navigation*, vol. 150, no. 5, pp. 356–365, 2003.
[10] Qualisys, http://www.qualisys.com/, 2015.
[11] R. C. Gonzalez and R. E. Woods, "Image processing," *Digital image processing*, vol. 2, 2007.
[12] N. Otsu, "A threshold selection method from gray-level histograms," *Automatica*, vol. 11, no. 285-296, pp. 23–27, 1975.